\documentclass[runningheads,a4paper]{llncs}

\usepackage{amsmath}
\usepackage{amssymb}
\usepackage{color}
\usepackage{sidecap}
\usepackage{mathtools}

\usepackage{amssymb}
\usepackage{graphicx}
\usepackage{wrapfig}

\usepackage{url}
\urldef{\mailsa}\path|{alfred.hofmann, ursula.barth, ingrid.haas, frank.holzwarth,|
\urldef{\mailsb}\path|anna.kramer, leonie.kunz, christine.reiss, nicole.sator,|
\urldef{\mailsc}\path|erika.siebert-cole, peter.strasser, lncs}@springer.com|    
\newcommand{\keywords}[1]{\par\addvspace\baselineskip
\noindent\keywordname\enspace\ignorespaces#1}

\newcommand{\bmx}[0]{\begin{bmatrix}}
\newcommand{\emx}[0]{\end{bmatrix}}

\newcommand{\qlay}[1]{\left[#1\right]}
\newcommand{\vect}[1]{\mathbf{#1}}
\newcommand{\vects}[1]{\boldsymbol{#1}}
\newcommand{\matr}[1]{\mathbf{#1}}

\newcommand{\vc}[0]{\vect{c}}
\newcommand{\vh}[0]{\vect{h}}

\newcommand{\vx}[0]{\vect{x}}
\newcommand{\vw}[0]{\vect{w}}

\newcommand{\vu}[0]{\vect{u}}

\newcommand{\mW}[0]{\matr{W}}

\newcommand{\mU}[0]{\matr{U}}
\newcommand{\mV}[0]{\matr{V}}

\newcommand{\TT}[0]{{\vects{\theta}}}

\newcommand{\RR}[0]{\mathbb{R}}
\newcommand{\Scal}[0]{\mathcal{S}}

\begin{document}

\mainmatter  

\title{Learned-Norm Pooling for Deep Feedforward and Recurrent Neural Networks}

\titlerunning{Learned-Norm Pooling for Deep Neural Networks}

%
%
\author{Caglar Gulcehre
\and Kyunghyun Cho 
\and Razvan Pascanu
\and Yoshua Bengio$^\star$}
%

\institute{D\'{e}partement d'Informatique et de Recherche Op\'{e}rationelle\\
Universit\'{e} de Montr\'{e}al \\
($\star$) CIFAR Fellow}

%
%

\maketitle

\begin{abstract}
    In this paper we propose and investigate a novel nonlinear
    unit, called $L_p$ unit, for deep neural networks.  The
    proposed $L_p$ unit receives signals from several projections
    of a subset of units in the layer below and computes a
    normalized $L_p$ norm.  We notice two interesting
    interpretations of the $L_p$ unit.  First, the proposed unit
    can be understood as a generalization of a number of
    conventional pooling operators such as average,
    root-mean-square and max pooling widely used in, for
    instance, convolutional neural networks (CNN), HMAX models
    and neocognitrons.  Furthermore, the $L_p$ unit is, to a
    certain degree, similar to the recently proposed maxout unit
    \cite{Goodfellow_maxout_2013} which achieved the
    state-of-the-art object recognition results on a number of
    benchmark datasets.  Secondly, we provide a geometrical
    interpretation of the activation function based on which we
    argue that the $L_p$ unit is more efficient at representing
    complex, nonlinear separating boundaries. Each $L_p$ unit
    defines a superelliptic boundary, with its exact shape
    defined by the order $p$. We claim that this makes it
    possible to model arbitrarily shaped, curved boundaries more
    efficiently by combining a few $L_p$ units of different
    orders. This insight justifies the need for learning
    different orders for each unit in the model. We empirically
    evaluate the proposed $L_p$ units on a number of datasets and
    show that multilayer perceptrons (MLP) consisting of the
    $L_p$ units achieve the state-of-the-art results on a number
    of benchmark datasets. Furthermore, we evaluate the proposed
    $L_p$ unit on the recently proposed deep recurrent neural
    networks (RNN).

\end{abstract}

\section{Introduction}

The importance of well-designed nonlinear activation functions
when building a deep neural network has become more apparent
recently.  Novel nonlinear activation functions that are
unbounded and often piecewise linear but not continuous such as
rectified linear units (ReLU)
\cite{Nair-2010,Glorot+al-AI-2011-small}, or rectifier, and
maxout units \cite{Goodfellow_maxout_2013} have been found to be
particularly well suited for deep neural networks on many object
recognition tasks. 

A pooling operator, an idea which dates back to the work in
\cite{Hubel+Wiesel-1968}, has been adopted in many object
recognizers. Convolutional neural networks which often employ max
pooling have achieved state-of-the-art recognition performances
on various benchmark datasets \cite{Krizhevsky-2012,Ciresan-et-al-2012}. Also, biologically
inspired models such as HMAX have employed max pooling
\cite{Riesenhuber+Poggio-1999}. A pooling operator, in this
context, is understood as a way to summarize a high-dimensional
collection of neural responses and produce features that
are invariant to some variations in the input (across the
filter outputs that are being pooled).

Recently, the authors of \cite{Goodfellow_maxout_2013} proposed to understand a
pooling operator itself as a nonlinear activation function. The
proposed maxout unit \textit{pools} a group of linear responses,
or outputs, of neurons, which overall acts as a piecewise linear
activation function.  This approach has achieved many
state-of-the-art results on various benchmark datasets.

In this paper, we attempt to generalize this approach by noticing
that most pooling operators including max pooling as well as
maxout units can be understood as special cases of computing a
normalized $L_p$ norm over the outputs of a set of filter outputs. 
Unlike those conventional pooling
operators, however, we claim here that it is beneficial to estimate
the order $p$ of the $L_p$ norm instead of fixing it to a certain
predefined value such as $\infty$, as in max pooling. 

The benefit of learning the order $p$, and thereby a neural
network with $L_p$ units of different orders, can be understood
from geometrical perspective. As each $L_p$ unit defines a
spherical shape in a non-Euclidean space whose metric is defined
by the $L_p$ norm, the combination of multiple such units leads
to a non-trivial separating boundary in the input space. 
In particular, an MLP may learn a highly curved boundary
efficiently by taking advantage of different values of $p$. In
contrast, using a more conventional nonlinear activation
function, such as the rectifier, results in boundaries that are
piece-wise linear. Approximating a curved separation of classes
would be more expensive in this case, in terms of the number of
hidden units or piece-wise linear segments.

In Sec.~\ref{sec:background} a basic description of a multi-layer
perceptron (MLP) is given followed by an explanation of how a
pooling operator may be considered a nonlinear activation
function in an MLP.  We propose a novel $L_p$ unit for an MLP by
generalizing pooling operators as $L_p$ norms in
Sec.~\ref{sec:lpunit}. In Sec.~\ref{sec:geometric} the proposed
$L_p$ unit is further analyzed from the geometrical perspective.
We describe how the proposed $L_p$ unit may be used by recurrent
neural networks in Sec.~\ref{sec:drnn}.  
Sec.~\ref{sec:exp} provides empirical evaluation of the $L_p$ unit 
on a number of object recognition tasks.

\section{Background}
\label{sec:background}

\subsection{Multi-layer Perceptron}

A multi-layer perceptron (MLP) is a feedforward neural network
consisting of multiple layers of nonlinear neurons \cite{Rosenblatt1962}.  Each neuron $u_j$ of an MLP typically
receives a weighted sum of the incoming signals $\left\{ a_1,
\dots, a_N \right\}$ and applies a nonlinear activation function
$\phi$ to generate a scalar output such that
\begin{align}
    \label{eq:usual_nonlin}
    u_j\left( \left\{ a_1, \dots, a_N \right\} \right) = \phi\left( \sum_{i=1}^N w_{ij} a_i
    \right).
\end{align}
With this definition of each neuron\footnote{We omit a bias to
make equations less cluttered.}, we define the output of an MLP
having $L$ hidden layers and $q$ output neurons given an input
$\vx$ by
\begin{align}
    \label{eq:mlp_output}
        \vu(\vx \mid \TT) =  \phi \left(\mU^\top \phi_{\qlay{L}} \left(
    \mW_{\qlay{L}}^\top 
    \cdots  
    \phi_{\qlay{1}} \left( \mW_{\qlay{1}}^\top \vx \right)
    \cdots \right) \right),
\end{align}
where $\mW_{\qlay{l}}$ and $\phi_{\qlay{l}}$ are the weights and
the nonlinear activation function of the $l$-th hidden layer, and
$\mW_{\qlay{1}}$ and $\mU$ are the weights associated with the
input and output, respectively.

\subsection{Pooling as a Nonlinear Unit in MLP}

Pooling operators have been widely used in convolutional neural
networks (CNN) \cite{LeCun+98,Fukushima80,Riesenhuber+Poggio-1999} to reduce
the dimensionality of a high-dimensional output of a
convolutional layer. 
When used to group spatially neighboring neurons, this operator
which summarizes a group of neurons in a lower layer is able to
achieve the property of (local) translation invariance. Various
types of pooling operator have been proposed and used
successfully, such as average pooling, root-of-mean-squared (RMS)
pooling and max pooling \cite{Jarrett-ICCV2009,Yang+al-2009}.

A pooling operator may be viewed instead as a nonlinear
activation function. It receives input signal from the layer
below, and it returns a scalar value. The output is the result of
applying some nonlinear function such as $\max$ (max pooling). 
The difference from traditional nonlinearities is that the
pooling operator is not applied element-wise on the lower layer,
but rather on groups of hidden units.  A \textit{maxout}
nonlinear activation function proposed recently in
\cite{Goodfellow_maxout_2013} is a representative example of
\textit{max} pooling in this respect. 

\begin{figure}[t]
    \centering
    \hfill
    \begin{minipage}{0.49\columnwidth}
        \centering
        \includegraphics[width=0.55\columnwidth]{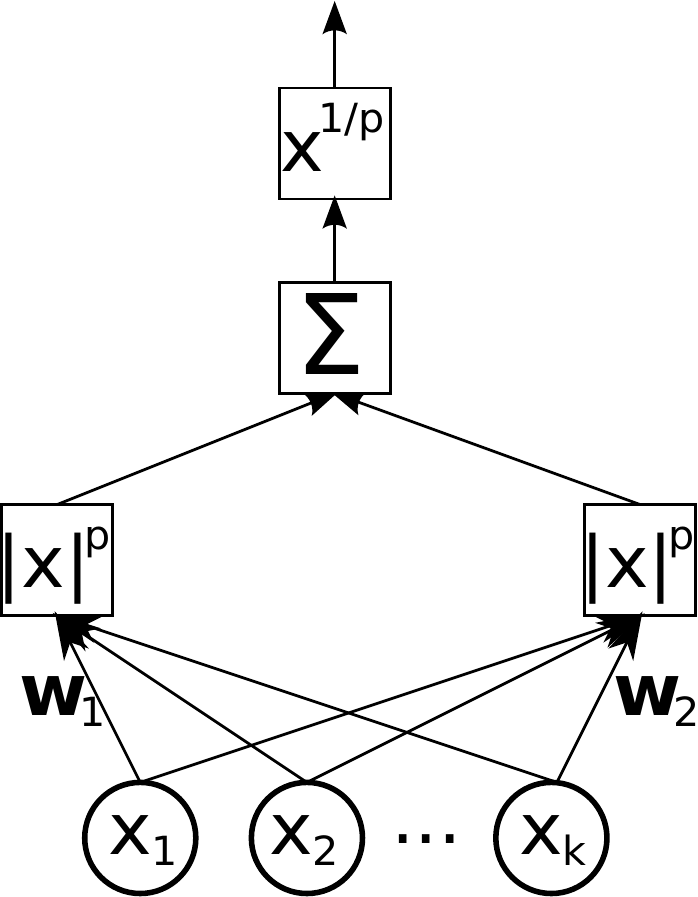}
        \\
        (a)
    \end{minipage}
    \hfill
    \begin{minipage}{0.49\columnwidth}
        \centering
        \includegraphics[width=0.99\columnwidth,clip=true,trim=50 20 50 20]{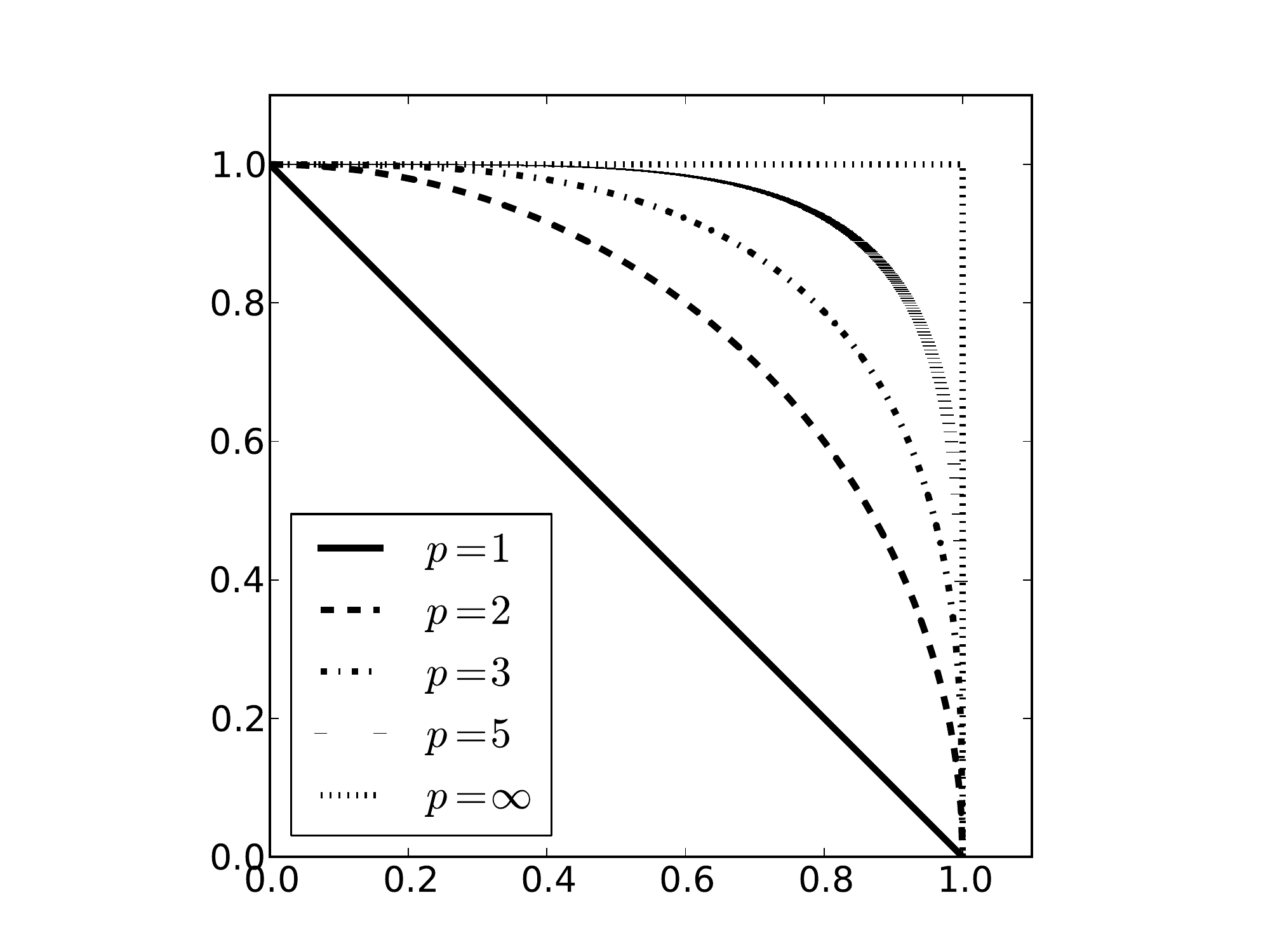}
        \\
        (b)
    \end{minipage}
    \caption{
        (a) An illustration of a single $L_p$ unit with two sets of incoming
        signals.  For clarity, biases and the division by the number of filters
        are omitted. The symbol $x$ in each block (square) represents an input signal to
        that specific block (square) only. (b) An illustration of the effect of $p$ on
        the shape of an ellipsoid. Only the first quadrant is shown. 
    }
     \label{figure:LpMLP}
     \vspace{-2mm}
\end{figure}

\section{$L_p$ Unit}
\label{sec:lpunit}

The recent success of maxout has motivated us to consider a more
general nonlinear activation function that is rooted in a pooling
operator. In this section, we propose and discuss a new nonlinear
activation function called an $L_p$ unit which replaces the
\textit{max} operator in a \textit{maxout} unit by an $L_p$ norm.

\subsection{Normalized $L_p$-norm}

Given a finite vector/set of input signals $\left[ a_1, \dots,
a_N \right]$ a normalized $L_p$ norm is defined as
\begin{align}
    \label{eq:lp_norm}
    u_j\left(\left[ a_1, \dots, a_N \right]\right)= \left( \frac{1}{N}
    \sum_{i=1}^N \left|a_i - c_i \right|^{p_j} \right)^{\frac{1}{p_j}},
\end{align}
where $p_j$ indicates that the order of the norm may differ for
each neuron. It should be noticed that when ${0 < p_j < 1}$ this
definition is not a norm anymore due to the violation of triangle
inequality. In practice, we re-parameterize $p_j$ by $1 +
\log\left( 1 + e^{\rho_j} \right)$ to satisfy this constraint.

The input signals (also called filter outputs) $a_i$ are defined
by
\[
    a_i = \vw_i^\top \vx, 
\]
where $\vx$ is a vector of activations from the lower layer.
$c_i$ is a center, or bias, of the $i$-th input signal $a_i$.
Both $p_j$ and $c_i$ are model parameters that are learned.

We call a neuron with this nonlinear activation function an $L_p$
unit.  An illustration of a single $L_p$ unit is presented in
Fig.~\ref{figure:LpMLP} (a).

Each $L_p$ unit in a single layer receives input signal from a
subset of linear projections of the activations of the layer
immediately below. In other words, we project the activations
of the layer immediately below linearly to ${A = \left\{a_1,
\dots, a_N \right\}}$. We then divide $A$ into equal-sized,
non-overlapping groups of which each is fed into a single $L_p$
unit. Equivalently, each $L_p$ unit has its private set
of filters.

The parameters of an MLP having one or more layers of $L_p$ units
can be estimated by using backpropagation \cite{Rumelhart86b},
and in particular we adapt the order $p$ of the
norm.\footnote{
    The activation function is continuous everywhere
    except a finite set of points, namely when $a_i - c_i$ is 0
    and the absolute value function becomes discontinuous.  We
    ignore these discontinuities, as it is done, for instance, in
    maxouts and rectifiers.
} In our experiments, we use Theano
\cite{bergstra+al:2010-scipy-short} to compute these partial
derivatives and update the orders $p_j$ (through the parametrization
of $p_j$ in terms of $\rho_j$), as usual with any other
parameters.

\subsection{Related Approaches}

Thanks to the definition of the proposed $L_p$ unit based on the
$L_p$ norm, it is straightforward to see that many previously
described nonlinear activation functions or pooling operators are
closely related to or special cases of the $L_p$ unit. Here we
discuss some of them.

When $p_j = 1$, Eq.~\eqref{eq:lp_norm} becomes
\[
    u_j\left(\left[ a_1, \dots, a_N \right]\right)= \frac{1}{N}
    \sum_{i=1}^N \left|a_i\right|.
\]
If we further assume $a_i \geq 0$, for instance, by using a
logistic sigmoid activation function on the projection of the
lower layer, the activation is reduced to computing the average
of these projections.  This is a form of average pooling,
where the non-linear projections represent the pooled layer. 
With a single filter, this is equivalent to the absolute
value rectification proposed in~\cite{Jarrett-ICCV2009}.
If
$p_j$ is $2$ instead of $1$, the root-of-mean-squared pooling
from \cite{Yang+al-2009} is recovered. 

As $p_j$ grows and ultimately approaches $\infty$, the $L_p$ norm becomes
\[
    \lim_{p_j \to \infty} u_j\left(\left[ a_1, \dots, a_N \right]\right) =
    \max \left\{ \left| a_1 \right|, \dots, \left| a_N \right| \right\}.
\]
When $N=2$, this is a generalization of a rectified linear unit
(ReLU) as well as the absolute value unit \cite{Jarrett-ICCV2009}.
If each $a_i$ is constrained to be non-negative, this corresponds
exactly to the maxout unit. 

In short, the proposed $L_p$ unit interpolates among different
pooling operators by the choice of its order $p_j$. This was
noticed earlier in \cite{boureau-icml-10} as well as
\cite{Yang+al-2009}. However, both of them stopped at analyzing
the $L_p$ norm as a pooling operator with a fixed order and
comparing those conventional pooling operators against each
other. The authors of \cite{bergstra+bengio+louradour:2011}
investigated a similar nonlinear activation function that was
inspired by the cells in the primary visual cortex. In
\cite{hyvarinen2007complex}, the possibility of learning $p$ has
been investigated in a probabilistic setting in computer vision.

On the other hand, in this paper, we claim that the order $p_j$
needs to, and can be \textit{learned}, just like all other
parameters of a deep neural network.  Furthermore, we conjecture
that (1) an optimal distribution of the orders of $L_p$ units
differs from one dataset to another, and (2) each $L_p$ unit in a
MLP requires a different order from the other $L_p$. These
properties also distinguish the proposed $L_p$ unit from the
conventional radial-basis function network~(see, e.g.,
\cite{Haykin-2008})

\begin{figure}[ht]
    \centering
    \begin{minipage}{0.33\textwidth}
        \centering
          \includegraphics[width=\textwidth]{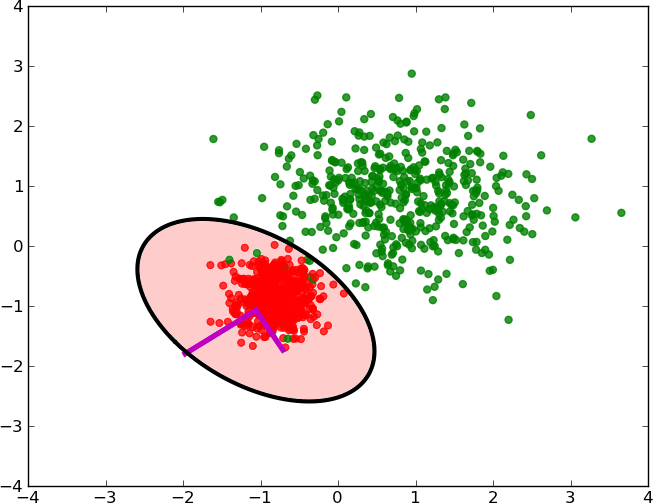}

          (a) $L_p$ with $p=2$
     \end{minipage}%
     \hfill
     \begin{minipage}{0.33\textwidth}
        \centering
          \includegraphics[width=\textwidth]{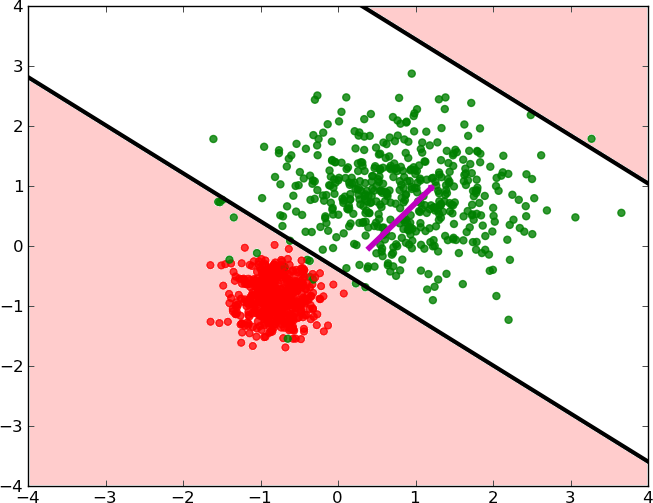}

          (b) $L_p$ with $p=\infty$
     \end{minipage}
     \hfill
    \begin{minipage}{0.33\textwidth}
        \centering
          \includegraphics[width=\textwidth]{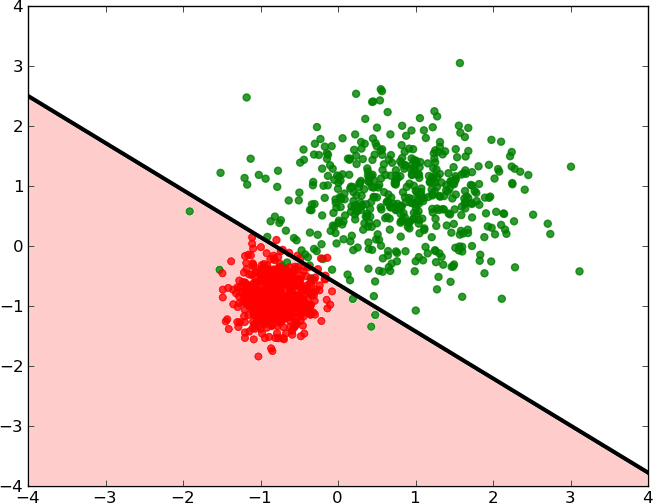}

          (c) Rectifier
    \end{minipage}

    \caption{Visualization of separating curves obtained using
        different activation functions. The underlying data
        distribution is a mixture of two Gaussian distributions.
        The red and green dots are the samples from the two
    classes, respectively, and the black curves are separating
curves found by the MLPs. The purple lines indicate the axes of
the subspace learned by each $L_p$ unit.  Best viewed in color.}
      \label{fig:twoGauss}
\end{figure}

\begin{figure}[ht]
        \centering
    \begin{minipage}{0.49\textwidth}
         \centering
          \includegraphics[width=0.8\textwidth]{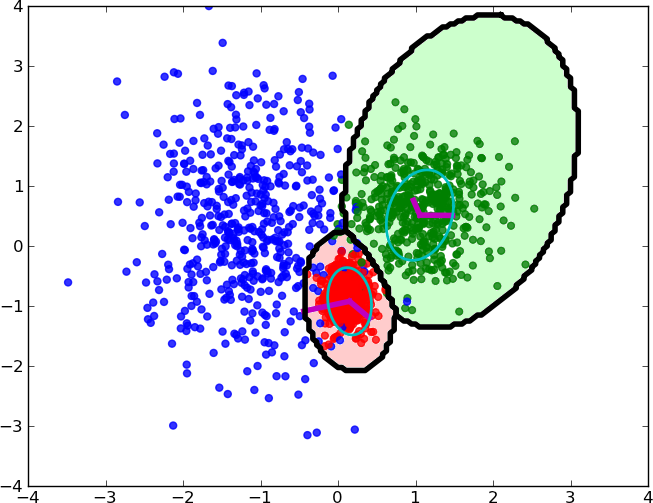}

          (a) $p_1 = p_2 = 2$
     \end{minipage}%
     \begin{minipage}{0.49\textwidth}
         \centering
          \includegraphics[width=0.8\textwidth]{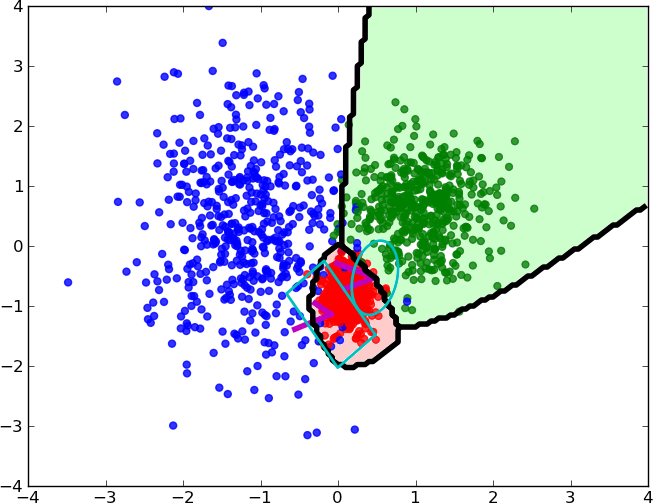}

          (b) $p_1 = 2$ and $p_2 = \infty$
     \end{minipage}
      \caption{Visualization of separating curves obtained using
          different orders of $L_p$ units. The underlying data
          distribution is a mixture of three Gaussian
          distributions.  The blue curves show the shape of the
          superellipse learned by each $L_p$ unit. The red, green
          and blue dots are the samples.
          Otherwise, the same color
          convention as in Fig.~\ref{fig:twoGauss} has been used.}
     \label{fig:threeGauss}
\end{figure}

\section{Geometrical Interpretation}
\label{sec:geometric}

We analyze the proposed $L_p$ unit from a geometrical perspective
in order to motivate our conjecture regarding the order of the
$L_p$ units. Let the value of an $L_p$ unit $u$ be given by:
\begin{align}
    \label{eq:single_lp}
    u(\vx) = \left(\frac{1}{N} \sum_{i=1}^N \left|\vw_i^\top \vx - c_i \right|^{p}\right)^{\frac{1}{p}},
\end{align}
where $\vw_i$ represents the $i$-th column of the matrix $\mW$.
The equation above effectively says that the $L_p$ unit computes
the $p$-th norm of the projection of the input $\vx$ on the
subspace spanned by $N$ vectors $\left\{ \vw_1, \dots, \vw_N
\right\}$. Let us further assume that $\vx \in \RR^d$ is a vector
in an Euclidean space. 

The space onto which $\vx$ is projected may be spanned by
linearly dependent vectors $\vw_i$'s. Due to the possible lack of
the linear independence among these vectors, they span a subspace
$\Scal$ of dimensionality $k \leq N$. The subspace $\Scal$ has
its origin at $\vc = \left[ c_1, \dots, c_N \right]$.

We impose a non-Euclidean geometry on this subspace by defining a
norm in the space to be $L_p$ with $p$ potentially not $2$, as in
Eq.~\eqref{eq:single_lp}. The geometrical object to which a
particular value of the $L_p$ unit corresponds forms a superellipse
when projected back into the original input space.
\footnote{
    Since $k \leq N$, the superellipse may be degenerate in the
    sense that in some of the $N - k$ axes the width may become
    infinitely large. However, as this does not invalidate our
    further argument, we continue to refer this kind of
    (degenerate) superellipse simply by an superellipse.
} 
The superellipse is centered at the inverse projection of $\vc$ in the Euclidean
input space.  Its shape  varies according to
the order $p$ of the $L_p$ unit and due to the potentially
linearly-dependent bases. As long as $p \geq 1$ the shape
remains convex.  Fig.~\ref{figure:LpMLP} (b) draws some of the
superellipses one can get with different orders of $p$,
as a function of $a_1$ (with a single filter).

In this way each $L_p$ unit partitions the input space into two
regions -- inside and outside the superellipse.  Each $L_p$ unit
uses a curved boundary of learned curvature to divide the space.
This is in contrast to, for instance, a maxout unit which uses
piecewise linear hyperplanes and might require more linear pieces
to approximate the same curved segment.  

\subsection{Qualitative Analysis in Low Dimension}

When the dimensionality of the input space is $2$ and each $L_p$
receives $2$ input signals, we can visualize the partitions of
the input space obtained using $L_p$ units as well as
conventional nonlinear activation functions. Here, we examine
some artificially generated cases in a $2$-D space. 

\subsubsection{Two Classes, Single $L_p$ Unit}

Fig.~\ref{fig:twoGauss} shows a case of having two classes
(\textcolor{red}{$\bullet$} and \textcolor{green}{$\bullet$}) of
which each corresponds to a Gaussian distribution. We trained
MLPs having a single hidden neuron. When the MLPs had an $L_p$
unit, we fixed $p$ to either $2$ or $\infty$. We can see in
Fig.~\ref{fig:twoGauss} (a) that the MLP with the $L_2$ unit
divides the input space into two regions -- inside and outside a
rotated superellipse.\footnote{
    Even though we use $p=2$, which means an Euclidean space, we
    get a superellipse instead of a circle because of the
    linearly-dependent bases $\left\{ \vw_1, \dots, \vw_N
    \right\}.$
}
The superellipse correctly identified one of the classes (red).

In the case of $p=\infty$, what we see is a degenerate rectangle
which is an extreme form of a superellipse. The superellipse
again spotted one of the classes and appropriately draws
a separating curve between the two classes.

In the case of rectifier units it could find a correct separating
curve, but it is clear that a single rectifier unit can only
partition the input space \textit{linearly} unlike $L_p$ units. A
combination of several rectifier units can result in a nonlinear
boundary, specifically a piecewise-linear one, though our claim is 
that you need more such rectifier 
units to get an arbitrarily shaped curve whose curvature changes
in a highly nonlinear way.

\subsubsection{Three Classes, Two $L_p$ Units}

Similarly to the previous experiment, we trained two MLPs having
two $L_p$ units on data generated from a mixture of three
Gaussian distribution.  Again, each mixture component corresponds
to each class. 

For one MLP we fixed the orders of the two $L_p$ units to $2$. In
this case, see Fig.~\ref{fig:threeGauss} (a), the separating
curves are constructed by combining two translated superellipses
represented by the $L_p$ units.  These units were able to locate
the two classes, which is sufficient for classifying the three
classes (\textcolor{red}{$\bullet$}, \textcolor{green}{$\bullet$}
and \textcolor{blue}{$\bullet$}).

The other MLP had two $L_p$ units with $p$ fixed to $2$ and
$\infty$, respectively. The $L_2$ unit defines, as usual, a
superellipse, while the $L_{\infty}$ unit defines a rectangle.
The separating curves are constructed as a combination of the
translated superellipse and rectangle and may have more
non-trivial curvature as in Fig.~\ref{fig:threeGauss} (b).

Furthermore, it is clear from the two plots in
Fig.~\ref{fig:threeGauss} that the curvature of the separating
curves may change over the input space. It will be easier to
model this non-stationary curvature using multiple $L_p$ units
with different $p$'s.

\begin{figure}[htp]
        \centering
    \begin{minipage}{0.48\columnwidth}
        \centering
          \includegraphics[width=\textwidth,clip=true,trim=90 50 80 60]{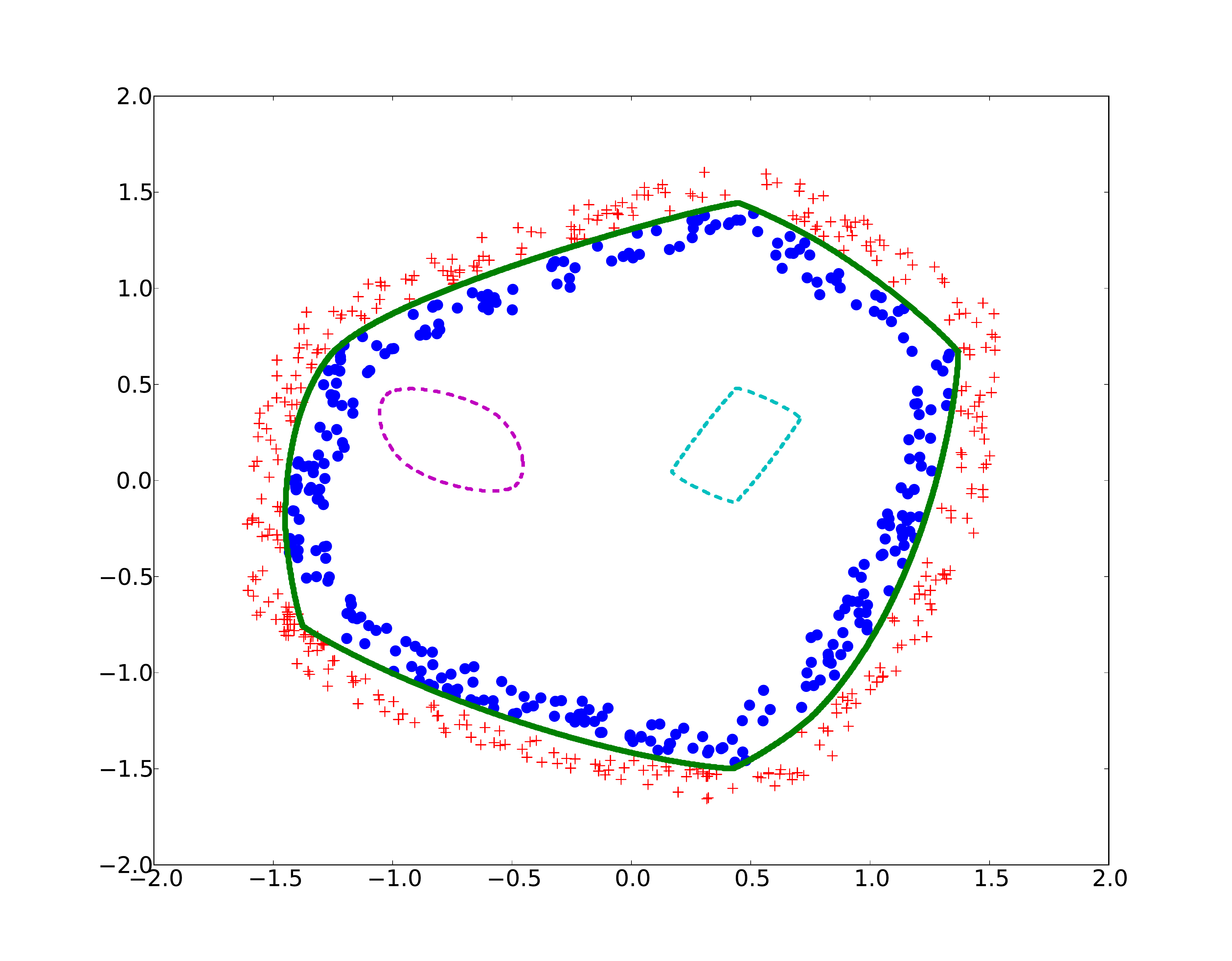}

          (a) 
     \end{minipage}
     \hfill
    \begin{minipage}{0.48\columnwidth}
        \centering
          \includegraphics[width=\textwidth,clip=true,trim=90 50 80 60]{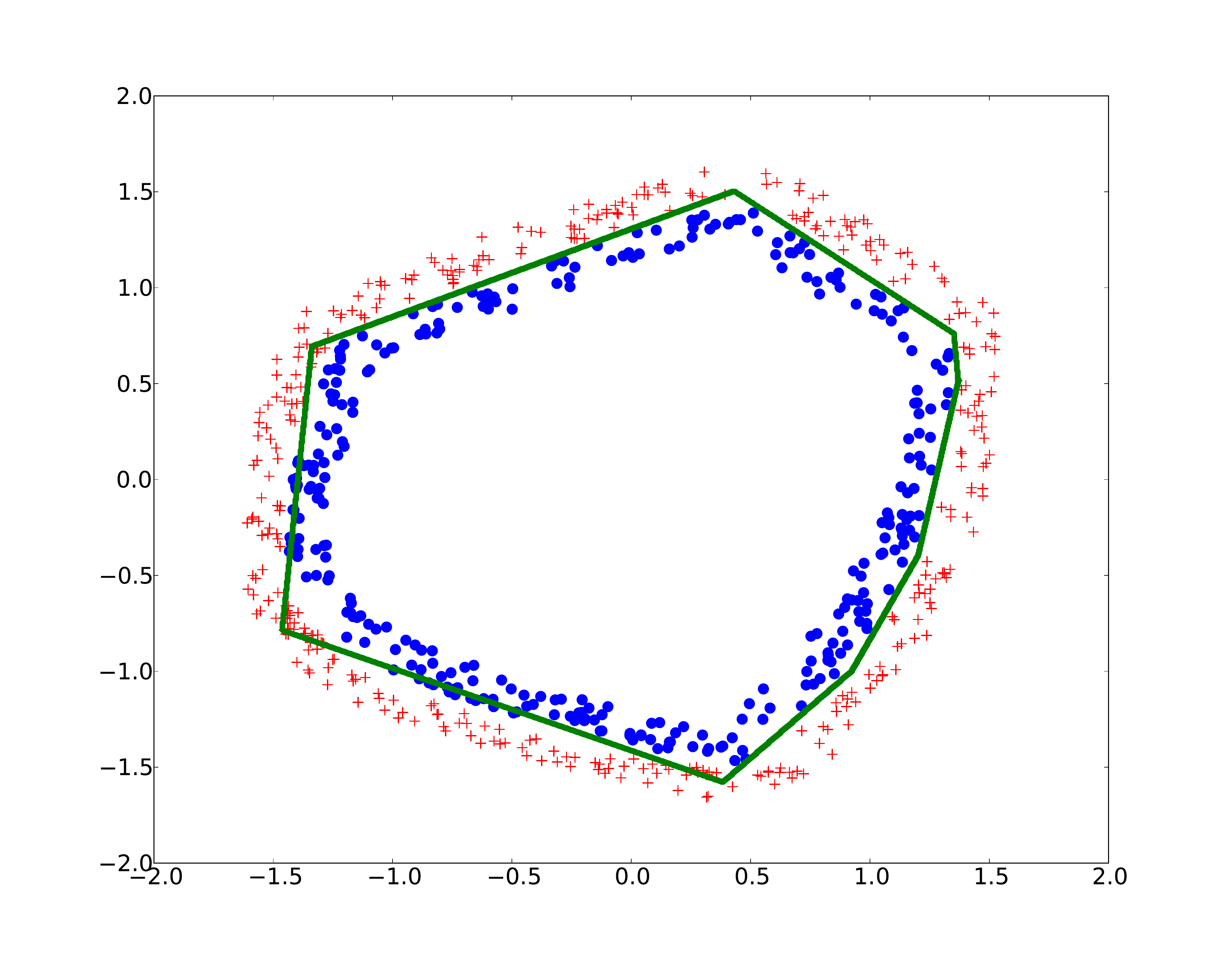}

          (b) 
     \end{minipage}

     \begin{minipage}{0.99\textwidth}
     \vspace{0pt}
     \centering
        \centering
          \includegraphics[width=\textwidth,clip=true,trim=0 0 100 50]{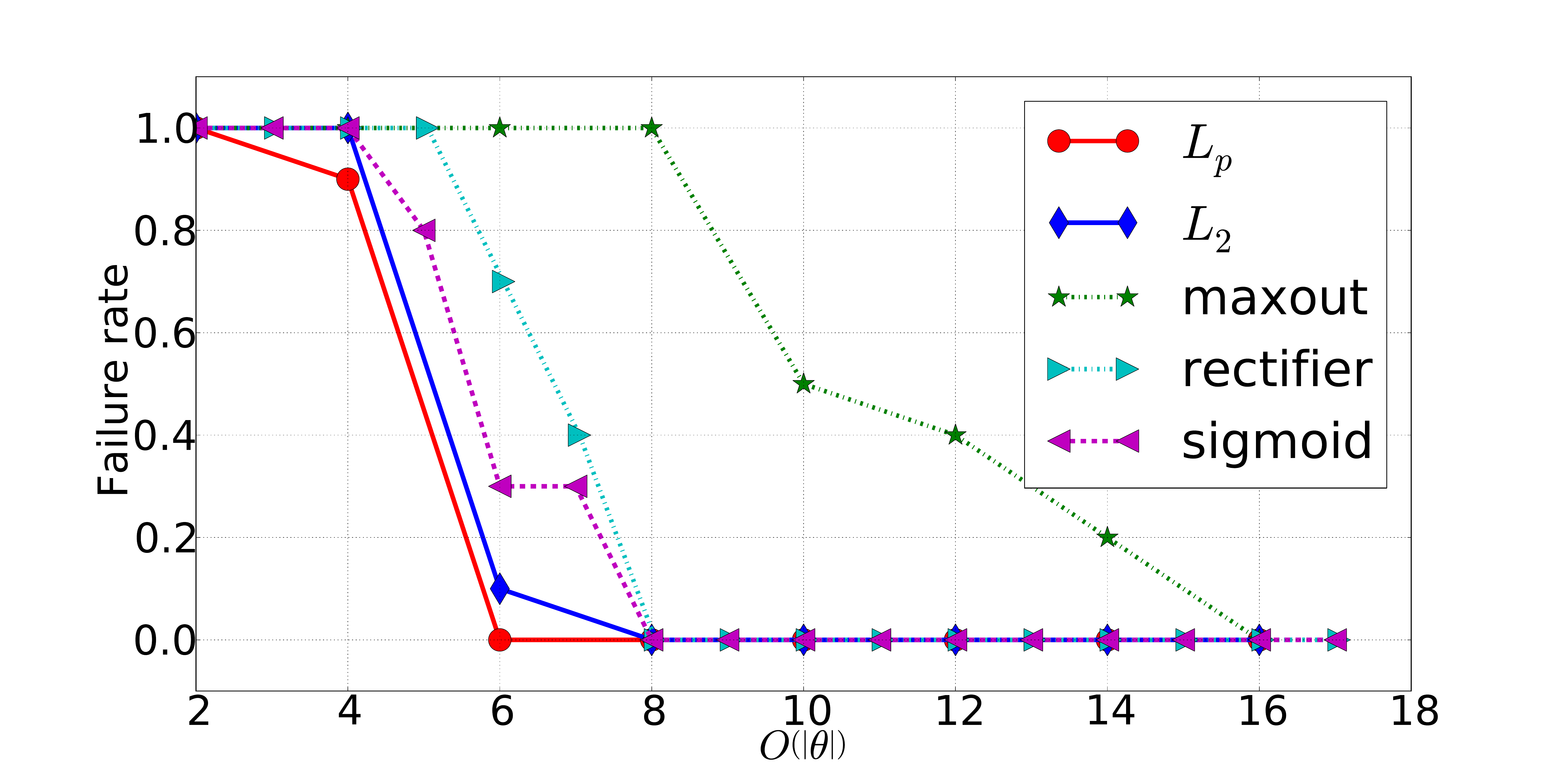}

          (c) 
     \end{minipage}

     \caption{(a) Visualization of data (two classes,
         \textcolor{red}{$+$} and \textcolor{blue}{$\bullet$}), a
         decision boundary learned by an MLP with two $L_p$ units
         (\textcolor{green}{green} curve) and the shapes
         corresponding to the orders $p$'s learned by the $L_p$
         units (\textcolor{magenta}{purple} and \textcolor{cyan}{cyan}
         dashed curves). (b) The same visualization done using four rectifiers.  
         (c) The failure rates computed with MLPs using different numbers of different
         nonlinear activation functions ($L_p$:
         \textcolor{red}{red} solid curve with \textcolor{red}{red $\bullet$}, $L_2$:
         \textcolor{blue}{blue} solid curve with \textcolor{blue}{blue $\blacklozenge$}, 
         maxout: \textcolor{green}{green} dashed curve with \textcolor{green}{green $\star$}, 
         rectifier:
         \textcolor{cyan}{cyan} dash-dot curve with \textcolor{cyan}{cyan $\blacktriangleright$} and 
         sigmoid: \textcolor{magenta}{purple} dashed curve with \textcolor{magenta}{purple $\blacktriangleleft$}). 
         The curves
         show the proportion of the failed attempts over ten random trials (y-axis) 
         against either the number of units for sigmoid and rectifier model or the total number of linear projection 
         going into the maxout units or $L_p$ units (x-axis). 
     }
     \label{fig:almostcircle}
\end{figure}

\begin{figure}[ht]
 \centering
 \begin{minipage}{0.32\textwidth}
     \centering
     \includegraphics[width=0.95\columnwidth]{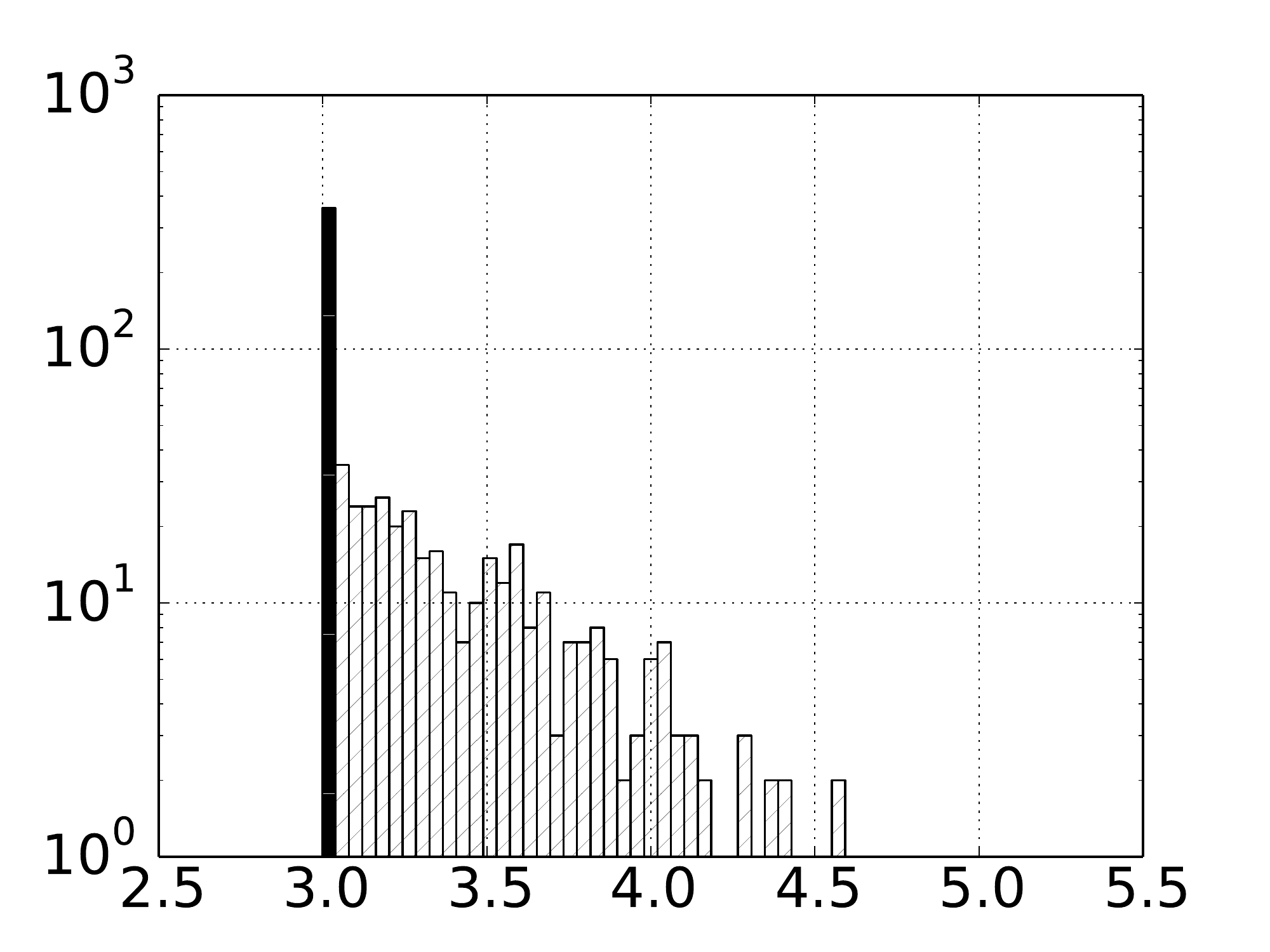}
     \\
     (a) MNIST
 \end{minipage}
 \hfill
 \begin{minipage}{0.32\textwidth}
     \centering
     \includegraphics[width=0.95\columnwidth]{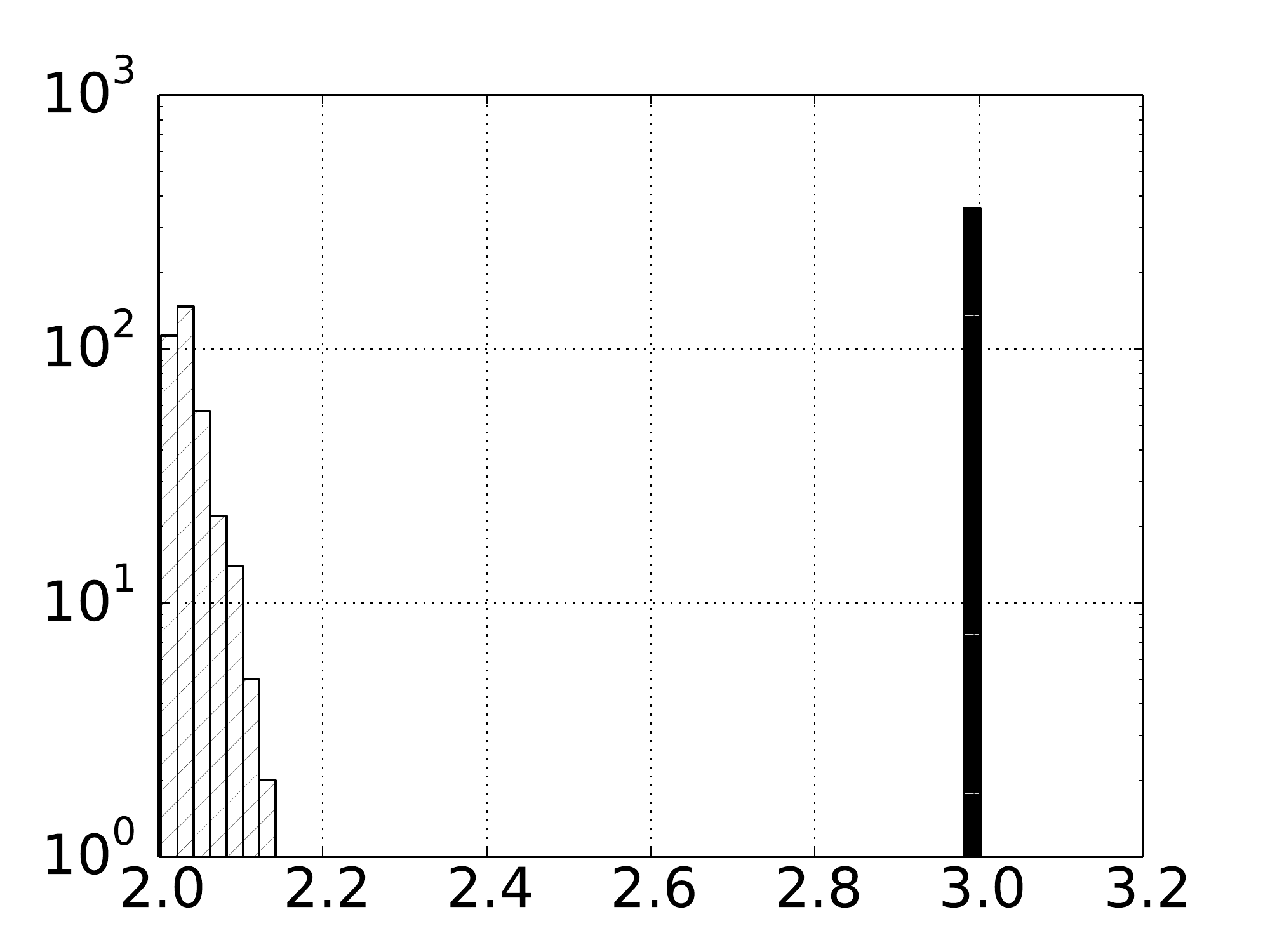}
     \\
     (b) TFD
 \end{minipage}
 \hfill
 \begin{minipage}{0.32\textwidth}
     \centering
     \includegraphics[width=0.95\columnwidth]{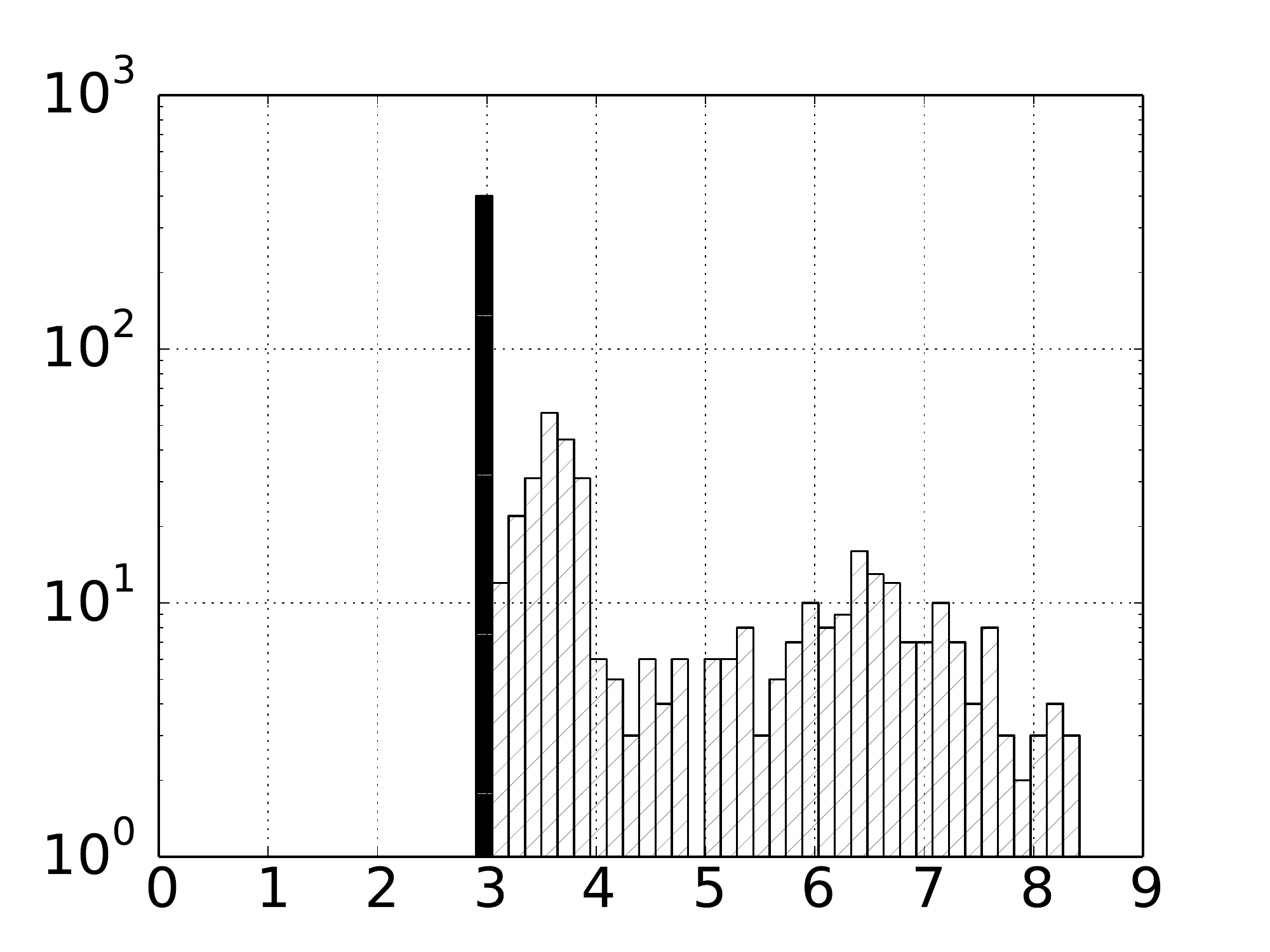}
     \\
     (c) Pentomino
 \end{minipage}

 \caption{Distributions of the initial (black bars $\blacksquare$) and
     learned (shaded bars $\diagup$) orders on MNIST, TFD and
     Pentomino. $x$-axis and $y$-axis show the order and the number of $L_p$
     units with the corresponding order. Note the difference in the scales of
 the x-axes and that the y-axes are in logarithmic scale.}
\label{fig:order_dist}
\end{figure}

\subsubsection{Decision Boundary
with Non-Stationary Curvature: Representational Efficiency}

In order to test the potential efficiency of the proposed $L_p$
unit from its ability to learn the order $p$, we have designed a
binary classification task that has a decision boundary with a
non-stationary curvature. We use 5000 data points of which a subset is shown in
Fig.~\ref{fig:almostcircle} (a), where two classes are marked
with blue dots (\textcolor{blue}{$\bullet$}) and red crosses (\textcolor{red}{$+$}),
respectively.

On this dataset, we have trained MLPs with either $L_p$ units,
$L_2$ units ($L_p$ units with fixed $p=2$), maxout units,
rectifiers or logistic sigmoid units. We varied the number of
parameters, which correspond to the number of units in the case
of rectifiers and logistic sigmoid units and to the number of
inputs signals to the hidden layer in the case of $L_p$ units,
$L_2$ units and maxout units, from $2$ to $16$. For each setting,
we trained ten randomly initialized MLPs. In order to reduce
effects due to optimization difficulties, we used in all cases
natural conjugate gradient \cite{Pascanu+Bengio-arxiv2013}.

From Fig.~\ref{fig:almostcircle} (c), it is clear that the MLPs
with $L_p$ units outperform all others in terms of representing 
this specific curve. They were able to achieve
the zero training error with only three units (i.e., 6 filters)
on all ten random runs and achieved the
lowest average training error even with less units.
Importantly, the comparison to the performance of the MLPs with
$L_2$ units shows that it is beneficial to learn the orders $p$
of $L_p$ units. For example, with only two $L_2$ units none of the 
ten random runs succeed while at least one succeeds with two
$L_p$ units. All the other MLPs, especially ones with
rectifiers and maxout units which can only model the decision
boundary with piecewise linear functions, were not able to
achieve the similar efficiency of the MLPs with $L_p$ units 
(see Fig~\ref{fig:almostcircle} (b)). 

Fig.~\ref{fig:almostcircle} (a) also shows the decision boundary
found by the MLP with two $L_p$ units after training. As can be
observed from the shapes of the $L_p$ units
(\textcolor{magenta}{purple} and \textcolor{cyan}{cyan} dashed
curves), each $L_p$ unit learned an appropriate order $p$ that
enables them to model the non-stationary decision boundary.
Fig.~\ref{fig:almostcircle} (b) shows the boundary obtained by 
a rectifier model with four units. We can see that it has to 
use linear segments to compose the boundary, resulting in not 
perfectly solving the task. The rectifier model represented here 
has 64 mistakes, versus 0 obtained by the $L_p$ model. 

Although this is a low-dimensional, artificially generated
example, it demonstrates that the proposed $L_p$ units
are efficient at representing decision boundaries which have
non-stationary curvatures.

\section{Application to Recurrent Neural Networks}
\label{sec:drnn}

A conventional recurrent neural network (RNN) mostly uses
saturating nonlinear activation functions such as $\tanh$ to
compute the hidden state at each time step. This prevents the
possible explosion of the activations of hidden states over time
and in general results in more stable learning dynamics.
However, at the same time, this does not allow us to build an RNN
with recently proposed non-saturating activation functions such
as rectifiers and maxout as well as the proposed $L_p$ units.

The authors of \cite{Pascanu2013} recently proposed three ways to extend the
conventional, \textit{shallow} RNN into a deep RNN. Among those
three proposals, we notice that it is possible to use
non-saturating activations functions for a deep RNN with deep
transition without causing the instability of the model, because
a saturating non-linearity ($\tanh$) is applied in sandwich between
the $L_p$ MLP associated with each step.

The deep transition RNN (DT-RNN) has one or more intermediate
layers between a pair of consecutive hidden states. The
transition from a hidden state $\vh_{t-1}$ at time $t-1$ to the
next hidden state $\vh_t$ is
\begin{align*}
    \label{eq:deeptrans}
    \vh_{t} = g\left( \mW^\top f\left( \mU^\top \vh_{t-1} +
    \mV^\top \vx_t \right) \right),
\end{align*}
not showing biases, as previously.

When a usual saturating nonlinear activation function is used for
$g$, the activations of the hidden state $\vh_t$ are bounded.
This allows us to use any, potentially non-saturating nonlinear
function for $f$. We can simply use a layer of the proposed $L_p$
unit in the place of $f$. 

As argued in \cite{Pascanu2013}, if the procedure of
constructing a new summary which corresponds to the new hidden
state $\vh_t$ from the combination of the current input $\vx_t$
and the previous summary $\vh_{t-1}$ is highly nonlinear, any
benefit of the proposed $L_p$ unit over the existing,
conventional activation functions in feedforward neural networks
should naturally translate to these deep RNNs as well. We show
this effect empirically later by training a deep output, deep
transition RNN (DOT-RNN) with the proposed $L_p$ units.

\section{Experiments}
\label{sec:exp}

In this section, we provide empirical evidences showing the
advantages of utilizing the $L_p$ units.  In order to clearly
distinguish the effect of employing $L_p$ units from introducing
data-specific model architectures, all the experiments in this
section are performed by neural networks having densely connected
hidden layers. 

\subsection{Claims to Verify}

Let us first list our claims about the proposed $L_p$ units that
need to be verified through a set of experiments. We expect the
following from adopting $L_p$ units in an MLP:

\begin{enumerate}
    \itemsep 0em
    \item The optimal orders of $L_p$ units vary across datasets
    \item An optimal distribution of the orders of $L_p$ units is not close
        to a (shifted) Dirac delta distribution
\end{enumerate}

The first claim states that there is no universally optimal order
$p_j$. We train MLPs on a number of benchmark datasets to see the
resulting distribution of $p_j$'s.  If the distributions had been
similar between tasks, claim 1 would be rejected.

This naturally connects to the second claim. As the orders are
estimated via learning, it is unlikely that the orders of all
$L_p$ units will convergence to a single value such as $\infty$
(maxout or max pooling), $1$ (average pooling) or $2$ (RMS
pooling).  We expect that the response of each $L_p$ unit will
specialize by using a distinct order. The inspection of the
trained MLPs to confirm the first claim will validate this claim
as well. 

On top of these claims, we expect that an MLP having $L_p$
units, when the parameters including the orders of the $L_p$
units are well estimated, will achieve highly competitive
classification performance.  In addition to classification tasks
using feedforward neural networks, we anticipate that a recurrent
neural network benefits from having $L_p$ units in the
intermediate layer between the consecutive hidden states, as well.

\begin{wraptable}{I}{0.4\textwidth}
    \centering
\begin{tabular}{c | c c}
\hline
Dataset    & Mean &  Std. Dev. \\
\hline
MNIST     & 3.44 & 0.38 \\
TFD       & 2.04 & 0.22 \\
Pentomino & 5.81 & 1.56 \\
\hline
\end{tabular}
\caption{The means and standard deviations of the estimated
orders of $L_p$ units.}
\label{tab:order_dist}
\end{wraptable}

\subsection{Datasets}

For feedforward neural networks or MLPs, we have used four
datasets; MNIST \cite{LeCun+98}, Pentomino
\cite{Gulcehre+Bengio-ICLR2013}, the Toronto Face Database (TFD)
\cite{Susskind2010} and Forest Covertype\footnote{We use
the first 16 principal components only.} (data split DS2-581)
\cite{Trebar2008}.  MNIST, TFD and Forest Covertype are three
representative benchmark datasets, and Pentomino is a relatively recently
proposed dataset that is known to induce a
difficult optimization challenge for a deep neural network. We
have used three music datasets from
\cite{Boulanger+al-ICML2012-small} for evaluating the effect of
$L_p$ units on deep recurrent neural networks.

\subsection{Distributions of the Orders of $L_p$ Units}

To understand how the estimated orders $p$ of the proposed $L_p$
unit are distributed we trained MLPs with a single $L_p$ layer on
MNIST, TFD and Pentomino.  We measured validation error to search for good hyperparameters,
including the number of $L_p$ units and number of filters (input signals) per
$L_p$ unit. However, for Pentomino,
we simply fixed the size of the $L_p$ layer to $400$, and each
$L_p$ unit received signals from six hidden units below.

In Table~\ref{tab:order_dist}, the averages and standard
deviations of the estimated orders of the $L_p$ units in the
single-layer MLPs are listed for MNIST, TFD and Pentomino. It is
clear that the distribution of the orders depend heavily on the
dataset, which confirms our first claim described earlier.  From
Fig.~\ref{fig:order_dist} we can clearly see that even in a
single model the estimated orders vary quite a lot, which
confirms our second claim. Interestingly, in the case of
Pentomino, the distribution of the orders consists of two
distinct modes. 

The plots in Fig.~\ref{fig:order_dist} clearly show that the
orders of the $L_p$ units change significantly from their initial
values over training. Although we initialized the orders of the
$L_p$ units around $3$ for all the datasets, the resulting
distributions of the orders are significantly different among
those three datasets. This further confirms both of our claims.
As a simple empirical confirmation we tried the same experiment
with the fixed $p=2$ on TFD and achieved a worse test error
of $0.21$.

\subsection{Generalization Performance}

The ultimate goal of any novel nonlinear activation function for
an MLP is to achieve better generalization performance. We
conjectured that by learning the orders of $L_p$ units an MLP with
$L_p$ layers will achieve highly competitive classification
performance.

For MNIST we trained an MLP having two $L_p$ layers followed by a
softmax output layer. We used a recently introduced
regularization technique called dropout
\cite{Hinton-et-al-arxiv2012}. With this MLP we were able to
achieve $99.03$\% accuracy on the test set, which is comparable to
the state-of-the-art accuracy of $99.06$\% obtained by the
MLP with maxout units \cite{Goodfellow_maxout_2013}. 

On TFD we used the same MLP from the previous experiment to
evaluate generalization performance. We achieved a
recognition rate of $79.25$\%. Although we use neither
pretraining nor unlabeled samples, our result is close to the
current state-of-the-art rate of $82.4$\% on the
permutation-invariant version of the task reported by
\cite{Ranzato2013} who pretrained their models with a large
amount of unlabeled samples. 

As we have used the five-fold cross validation to find the
optimal hyperparameters,
we were able to use this to investigate the variance of the
estimations of the $p$ values. Table~\ref{tab:order_dist_TFD}
shows the averages and standard deviations of the estimated
orders for MLPs trained on the five folds using the best
hyperparameters. It is clear that in all the cases the orders
ended up in a similar region near two without too much
difference in the variance. 

Similarly, we have trained five randomly initialized MLPs on
MNIST and observed the similar phenomenon of all the resulting
MLPs having similar distributions of the orders. The standard
deviation of the averages of
the learned orders was only $0.028$, while its mean is
$2.16$.

The MLP having a single $L_p$ layer was able to classify the test
samples of Pentomino with $31.38$\% error rate. This is the best
result reported so far on Pentomino dataset
\cite{Gulcehre+Bengio-ICLR2013} without using any kind of prior
information about the task (the best previous result was 44.6\%
error). 

On Forest Covertype an MLP having three $L_p$ layers was trained.
The MLP was able to classify the test samples with only $2.83$\%
error.  The improvement is large compared to the previous
state-of-the-art rate of $3.13$\% achieved by the manifold
tangent classifier having four hidden layers of logistic sigmoid
units \cite{Dauphin-et-al-NIPS2011-small}. The result obtained
with the $L_p$ is comparable to that obtained with the MLP having
maxout units.

\begin{table}[t]
    \centering
    \begin{tabular}{r | c c c c c}
        \hline
        Data & MNIST & TFD & Pentomino & Forest Covertype \\
        \hline
        \hline
        $L_p$ & 
                $0.97$ \% & 
                $20.75$ \% &
                $31.85$ \% &
                $2.83$ \% \\
        \hline
        Previous & 
                    $0.94$ \%\footnotemark[1] & 
                    $21.29$ \%\footnotemark[2] &
                   $44.6$ \%\footnotemark[3] &
                   $2.78$ \%\footnotemark[4] \\
        \hline
    \end{tabular}
    \caption{The generalization errors on three datasets obtained
        by MLPs using the proposed $L_p$ units. The previous state-of-the-art
    results obtained by others are also presented for comparison.}
    \label{tab:overall_result}
\end{table}

\footnotetext[1]{Reported in
    \cite{Goodfellow_maxout_2013}.}
\footnotetext[2]{
    This result was obtained by training an MLP with rectified
    linear units which outperformed an MLP with maxout units. }
\footnotetext[3]{Reported in
    \cite{Gulcehre+Bengio-ICLR2013}.}
\footnotetext[4]{
    This result was obtained by training an MLP with maxout units which
    outperformed an MLP with rectified linear units.
}

\begin{wraptable}{I}{0.4\textwidth}
    \centering
\begin{tabular}{c | c c}
\hline
Fold    & Mean &  Std. Dev. \\
\hline
1 & 2.00 & $0.24$  {\tiny $\times 10^{-4}$} \\
2 & 2.00 & $0.24$  {\tiny $\times 10^{-4}$} \\
3 & 2.01 & $0.77$  {\tiny $\times 10^{-4}$} \\
4 & 2.02 & $1.50$ {\tiny $\times 10^{-4}$} \\
5 & 2.00 & $0.24$  {\tiny $\times 10^{-4}$} \\
\hline
\end{tabular}

\caption{The means and standard deviations of the estimated
orders of $L_p$ units obtained during the hyperparameter search
using the 5-fold cross-validation.}
\label{tab:order_dist_TFD}
\end{wraptable}

These results as well as previous best results for all datasets
are summarized in Table~\ref{tab:overall_result}.

In all experiments, we optimized hyperparameters such as an
initial learning rate and its scheduling to minimize validation
error, using random search
\cite{Bergstra+Bengio-2012-small}, which is generally more efficient
than grid search when the number of hyperparameters is not tiny. 
Each MLP was trained by
stochastic gradient descent. All the experiments
in this paper were done using the Pylearn2 library
\cite{pylearn2_arxiv_2013}. 

\subsection{Deep Recurrent Neural Networks}

\begin{wrapfigure}{L}{0.4\textwidth}
    \centering
    \includegraphics[width=0.38\textwidth]{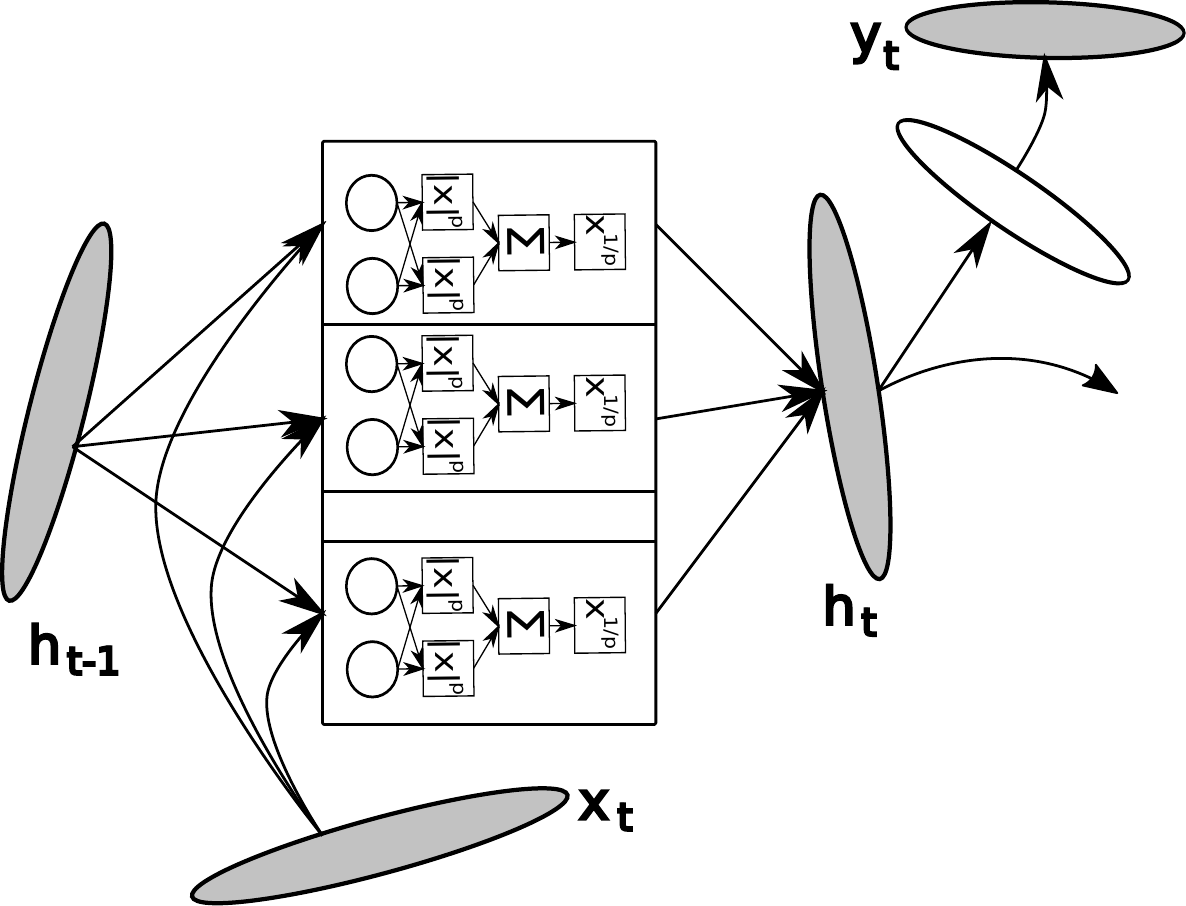}
    \caption{The illustration of the DOT-RNN using $L_p$ units.}
    \label{fig:dotrnn_lp}
\end{wrapfigure}

We tried the polyphonic music prediction tasks with three music datasets;
Nottingam, JSB and MuseData \cite{Boulanger+al-ICML2012-small}. The DOT-RNNs we
trained had deep transition with $L_p$ units and $\tanh$ units and deep output
function with maxout in the intermediate layer (see Fig.~\ref{fig:dotrnn_lp}
for the illustration). We coarsely optimized the size of the models and the
initial leaning rate as well as its schedule to maximize the performance on
validation sets. Also, we chose whether to threshold the norm of the gradient
based on the validation performance \cite{Pascanu+al-ICML2013-small}. All the
models were trained with dropout \cite{Hinton-et-al-arxiv2012}.

As shown in Table~\ref{tab:dotrnn_result}, we were able to achieve
the state-of-the-art results (RNN-only case) on all the three
datasets. These results are much better than those achieved by
the same DOT-RNNs using logistic sigmoid units in both deep
transition and deep output, which suggests the superiority of the
proposed $L_p$ units over the conventional saturating activation
functions. This suggests that the proposed $L_p$ units are well
suited not only to feedforward neural networks, but also to
recurrent neural networks. However, we acknowledge that more
investigation into applying $L_p$ units is needed in the future
to draw more concrete conclusion on the benefits of the $L_p$
units in recurrent neural networks.

\begin{table}[ht]
    \begin{minipage}{0.48\textwidth}
    \centering
    \begin{tabular}{c || c c |  c}
        \hline
        & \multicolumn{2}{c|}{DOT-RNN} & RNN \\
        Dataset & $L_p$ & sigmoid$^\star$ & *  \\
        \hline
        \hline
        Nottingam & 2.95 & 3.22 & 3.09 \\
        JSB & 7.92 & 8.44 & 8.01 \\
        Muse & 6.59 & 6.97 & 6.75 \\
    \end{tabular}
\end{minipage}
\hfill
\begin{minipage}{0.48\textwidth}
    \caption{
        The negative log-probability of the test sets computed by
        the trained DOT-RNNs.  ($\star$) These are the results
        achieved using DOT-RNNs having logistic sigmoid units,
        which we reported in \cite{Pascanu2013}.  (*) These are
        the previous best results achieved using conventional
        RNNs obtained in \cite{Bayer2013}.
    }
    \label{tab:dotrnn_result}
\end{minipage}
\vspace{-5mm}
\end{table}

\section{Conclusion}
\label{sec:conclusion}
In this paper, we have proposed a novel nonlinear activation
function based on the generalization of widely used pooling
operators. The proposed nonlinear activation function computes
the $L_p$ norm of several projections of the lower layer.  Max-,
average- and root-of-mean-squared pooling operators are special
cases of the proposed activation function, and naturally the
recently proposed maxout unit is closely related under an
assumption of non-negative input signals.

An important difference of the  $L_p$ unit from conventional
pooling operators is that the order of the unit is
\textit{learned} rather than pre-defined. We claimed that this
estimation of the orders is important and that the optimal model
should have $L_p$ units with various orders.

Our analysis has shown that an $L_p$ unit defines a non-Euclidean
subspace whose metric is defined by the $L_p$ norm. When
projected back into the input space, the $L_p$ unit defines an
ellipsoidal boundary. We conjectured and showed in a small scale
experiment that the combination of these curved boundaries may
more efficiently model separating curves of data with
non-stationary curvature.

These claims were empirically verified via training both deep
feedforward neural networks and deep recurrent neural networks.
We tested the feedforward neural network on on four benchmark
datasets; MNIST, Toronto Face Database, Pentomino and Forest
Covertype, and tested the recurrent neural networks on the task
of polyphonic music prediction. The experiments revealed that the
distribution of the estimated orders of $L_p$ units indeed
depends highly on dataset and is far away from a Dirac delta
distribution. Additionally, our conjecture that deep neural
networks with $L_p$ units will be able to achieve competitive
generalization performance was empirically confirmed.

\section*{Acknowledgments}
We would like to thank the developers of Pylearn2~\cite{pylearn2_arxiv_2013}
and Theano~\cite{bergstra+al:2010-scipy,Bastien-Theano-2012}.  We would also
like to thank CIFAR, and Canada Research Chairs for funding, and Compute
Canada, and Calcul Qu\'ebec for providing computational resources.

\bibliography{strings,ml,aigaion,myref}
\bibliographystyle{abbrv}

\end{document}